\documentclass[letterpaper, 10 pt, conference]{ieeeconf}  

\IEEEoverridecommandlockouts                              

\usepackage{colortbl}
\usepackage{graphicx}
\usepackage{booktabs}
\usepackage{hyperref}
\graphicspath{{Figures/}}
\usepackage{amsmath, float}
\usepackage{amssymb,bbm, array}
\usepackage[table,xcdraw]{xcolor}
\usepackage{multirow}
\title{\LARGE \bf
An Unsupervised Domain Adaptive Approach for Multimodal 2D Object Detection in Adverse Weather Conditions
}

\author{George Eskandar$^{1\star}$, Robert A. Marsden$^{1\star}$, Pavithran Pandiyan$^{1}$, Mario D\"obler$^{1}$, Karim Guirguis$^{2}$ and Bin Yang$^{1}$
\thanks{$^{\star}$ both authors contributed equally to this work}
\thanks{$^{1}$ University of Stuttgart, Institute of Signal Processing and System Theory, Stuttgart, Germany. {\tt\small \{george.eskandar, robert.marsden, mario.doebler, bin.yang\}@iss.uni-stuttgart.de}
}
\thanks{$^{2}$ Robert Bosch GmbH, Renningen, Germany. {\tt\small \{karim.guirguis\}@de.bosch.com}}
}

\begin{document}

\maketitle
\thispagestyle{empty}
\pagestyle{empty}
\begin{abstract}
Integrating different representations from complementary sensing modalities is crucial for robust scene interpretation in autonomous driving. While deep learning architectures that fuse vision and range data for 2D object detection have thrived in recent years, the corresponding modalities can degrade in adverse weather or lighting conditions, ultimately leading to a drop in performance. Although domain adaptation methods attempt to bridge the domain gap between source and target domains, they do not readily extend to heterogeneous data distributions. In this work, we propose an unsupervised domain adaptation framework, which adapts a 2D object detector for RGB and lidar sensors to one or more target domains featuring adverse weather conditions. Our proposed approach consists of three components. First, a data augmentation scheme that simulates weather distortions is devised to add domain confusion and prevent overfitting on the source data. Second, to promote cross-domain foreground object alignment, we leverage the complementary features of multiple modalities through a multi-scale entropy-weighted domain discriminator. Finally, we use carefully designed pretext tasks to learn a more robust representation of the target domain data. Experiments performed on the DENSE dataset show that our method can substantially alleviate the domain gap under the single-target domain adaptation (STDA) setting and the less explored yet more general multi-target domain adaptation (MTDA) setting. 
\end{abstract}

\section{Introduction}
In recent years, autonomous driving has made great progress due to the advances in deep learning, parallel computing, and the availability of large datasets. A critical component required for safe navigation is the perception tasks that extract information from the environment. Since these tasks are located at the beginning of autonomous driving pipelines, they should satisfy three conditions: accuracy, robustness, and real-time capability. In particular, the first two properties can benefit from the integration of different sensors, such as RGB cameras and active range sensors, like lidar and radar. The reason is that these sensors provide complementary information about the surroundings of the ego vehicle. While cameras can perceive the color, texture, and detailed shape of various objects, range sensors, in particular lidar, deliver 3D geometric information about the environment in the form of point clouds. The advantage of point clouds is that the representation of objects may be less affected by different textures, colors or lighting conditions. Therefore, fusing the information of lidar and camera is frequently used in perception tasks such as 2D and 3D object detection (OD) \cite{asvadi2017multimodal, schlosser2016fusing, liang2019multi, wang2019frustum, pfeuffer2018optimal, kim2018robust, wang2017fusing}. However, although RGB and lidar are complementary, they both degrade in adverse weather conditions. 

Multimodal deep learning networks which were trained to minimize
the empirical risk on a training dataset, usually exhibit poor generalization to data originating from a different domain. If the corresponding training dataset does not account for adverse weather conditions, the performance of the model typically decreases when evaluated in such a scenario. This paper addresses the problem of asymmetric degradation of RGB and lidar fusion models by adverse environmental conditions as an unsupervised domain adaptation (UDA) problem. 

UDA is a learning paradigm that seeks to solve the problem of poor generalization to out-of-domain data by transferring knowledge from a labeled source domain to an unlabeled target domain. UDA methods can be applied to a single modality, like RGB images \cite{iros_grasping_DA, iros_weatherDA_freespacesegmentation} and lidar \cite{iros_lidar_DA, icra_lidarnet}, or less commonly to multiple modalities \cite{multiactionrecognition, iros_heatnet_teacherstudent}. The latter setting is harder because the domain discrepancies manifest themselves differently for each modality. For instance, in 2D object detection, different lighting conditions affect RGB cameras by obscuring or blurring some parts of the image leading to undetected foreground objects. On the other hand, lidar remains reliable in these conditions but is vulnerable to rain and fog which lead to backscatter: Laser beams are reflected back from suspended water droplets in the air, which leads to formation of clutter in the resulting point cloud. Although noise filtering techniques \cite{drorfilter} can be deployed to remove some of the uncertain measurements, the resulting point cloud is still sparser and less informative. 



In this work, we present the first multimodal unsupervised domain adaptation framework for 2D object detection in autonomous driving using RGB and lidar. Starting from the domain adaptation (DA) theory \cite{36364}, our goal is to train a model that maps the source and target multimodal data to a common feature space which has the following properties: source domain feature discriminability, domain-invariant representations for class-specific features, and target domain feature discriminability. First, upon observing that the domain gap is manifested differently in RGB and lidar, we leverage multiple data augmentation techniques for the source domain to bring the domains closer together. Furthermore, we introduce a multi-scale discriminator with a feature matching loss and a maximum entropy fusion strategy. This helps to align foreground regions while effectively suppressing irrelevant background features. Finally, we exploit self-supervised learning (SSL), where we propose new variants of the rotation and the jigsaw puzzle pretext tasks. SSL is applied to both source and target domain data to enforce a domain-independent feature representation learning. We evaluate the proposed method on different weather splits of the DENSE dataset \cite{bijelic2019seeing} under two settings: single and multi-target UDA. Experiments show that the proposed method can effectively reduce the domain gap between the source and the target domain(s).



\section{Related Works}
\textbf{Multimodal tasks in Computer Vision}. Over the last decade, there has been many works on multimodal 2D object detection \cite{du2017car,schneider2017multimodal,mees2016choosing,takumi2017multispectral,guan2018fusion,pfeuffer2018optimal}. These works explore how and when to fuse features from different modalities. Fusion can occur by adding, concatenating, or ensembling features from different modalities and can take place at various depths of the network. More advanced methods use a mixture-of-experts \cite{mees2016choosing} or entropy fusion \cite{bijelic2019seeing} to weight the contribution of each modality proportionally to its information content. In \cite{bijelic2019seeing}, it was also shown that fusing different sensors can improve the generalization to unseen target domains. However, our experiments demonstrate that there still exists a significant domain gap. 

\textbf{Unsupervised Domain Adaptation}. UDA has been studied for perception tasks like classification, semantic segmentation, and object detection \cite{ADDA, CyCADACA, dafasterrcnn}. Typical approaches for domain adaptive object detection include pixel-level alignment using unpaired image-to-image translation \cite{CyCADACA, Hsu2020ProgressiveDA,  arruda2019ijcnn, Kim2019DiversifyAM} and feature-level alignment \cite{ADDA, dafasterrcnn, He2019MultiAdversarialFF, 10.1007/978-3-030-58621-8_24, 10.1007/978-3-030-58568-6_45, Ganin2015UnsupervisedDA}. Most of the approaches were applied to two-stage object detectors, like Faster R-CNN \cite{Ren2015FasterRT}, and do not handle multimodal data. Moreover, pixel-level alignment techniques based on CycleGAN \cite{Zhu2017UnpairedIT}, for example, do not generalize well to lidar because generative models do not extend to sparse data. Recently, multimodal domain adaptation (MMDA) has been investigated for tasks like video action recognition using images and optical flow \cite{multiactionrecognition, mmda_progressive}, object recognition using RGB and dense depth maps \cite{mmda_progressive}, RGB and text \cite{ mmda_image_text} and emotion recognition using visual and acoustic data \cite{mmda_emotion_recognition}. It has been shown that multimodal DA can be more efficient than single modality DA. These techniques are dependent on the task, modality, and network architecture and cannot be compared together. For instance, MMDA techniques can been applied on each modality alone before the fusion \cite{multiactionrecognition}, on the post-fusion features \cite{mmda_emotion_recognition}, or on both pre and post-fusion features \cite{multiactionrecognition, mmda_progressive, mmda_emotion_recognition}, according to the model design. In this work, we design a multimodal UDA approach for real-time one-stage 2D object detectors like YOLO \cite{Redmon2018YOLOv3AI} using RGB and sparse lidar depth maps. 

\textbf{Self-supervised and Representation Learning}. Self-supervised learning aims to learn a generalizable feature representation without requiring labels. To do this, it leverages various data augmentations or transformations on the unlabeled images and then trains a feature extractor to solve a so-called pretext task. There are many paradigms for how to learn a pretext task: the network can either predict the transform \cite{gidaris2018unsupervised,Noroozi2016UnsupervisedLO}, undo the transform like in image inpainting \cite{pathakCVPR16context}, or learn instance discrimination through a contrastive loss \cite{Misra2020SelfSupervisedLO}. SSL is promising for UDA because the features learned on unlabeled target domain data can be discriminative enough for other visual tasks like object detection and segmentation \cite{Xu2019SelfSupervisedDA}. In this work, we redesign and leverage some pretext tasks for domain adaptive OD.

\section{Method}
\textbf{Definitions}. Let $\mathcal{D}_{s} = \{ (x_{s}^{i},d_{s}^{i}, y_{s}^i)|_{i=1}^{N_s} \}$ denote a set of $N_s$ labeled source data pairs, where $x_s^i$ is an RGB image, $d_s^i$ is a projected lidar point cloud into the camera space and $y_s^i$ contains the instance labels, which consist of a bounding box and a class label for each object instance. In single-target UDA, we try to transfer knowledge from $\mathcal{D}_{s}$ to an unlabeled target domain $\mathcal{D}_t = \{ (x_t^i,d_t^i)|_{i=1}^{N_t} \}$, whereas in multi-target UDA, we aim to transfer knowledge to $m$ unlabeled target domains ($\mathcal{D}_{t_1}, \mathcal{D}_{t_2}, ..., \mathcal{D}_{t_m}$). An example for the latter case is \textit{Clear Day} $\rightarrow$ \{\textit{Snow}, \textit{Dense Fog}, ..., \textit{Night}\}.

\begin{figure*}[t!]
    \centering
    \includegraphics[width=1.0\textwidth, height=0.3 \textwidth]{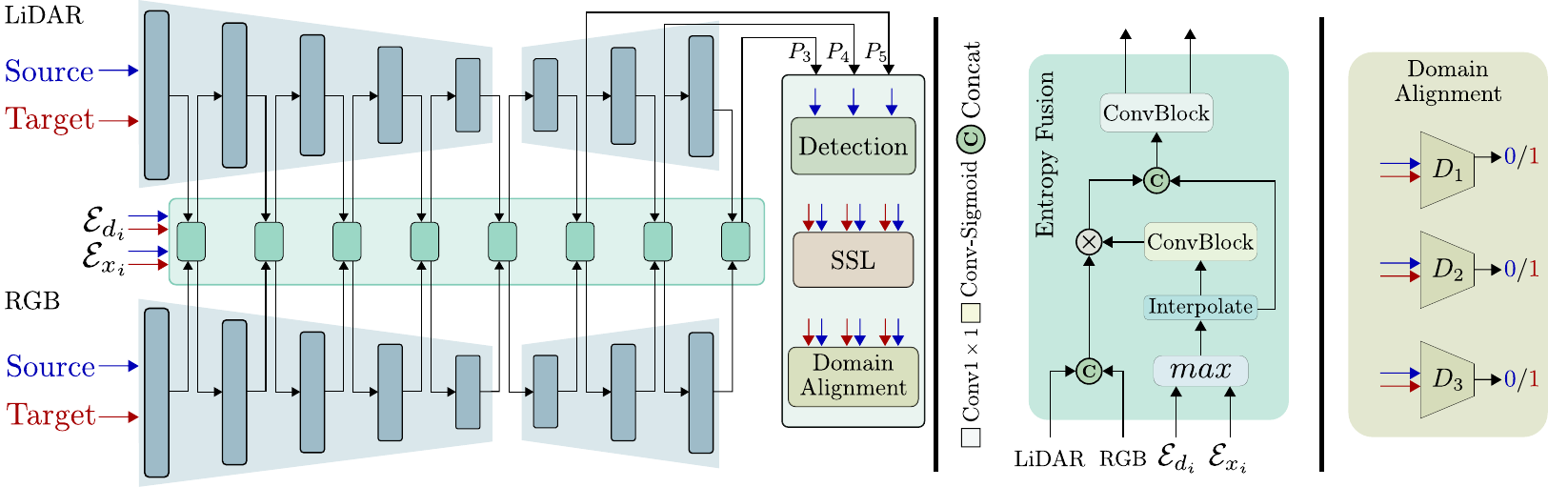}
    \vspace{-1em}
    \caption{Abstract illustration of our YOLO-V3 based network that fuses features of separate lidar and RGB branches with entropy fusion modules. The last \textit{ConvBlock} in each entropy module consists of a conv ($k=1 \times 1$), batchnorm and ReLU layer and reduces the number of input channels by 2.}
    \label{fig:architecture}
\end{figure*}

\textbf{Network}. Our work builds upon the entropy-steered deep fusion approach of \cite{bijelic2019seeing}. However, unlike \cite{bijelic2019seeing}, we use the well-known YOLO-V3 \cite{Redmon2018YOLOv3AI} detection network, which we adapt as follows. As shown in Fig. \ref{fig:architecture} schematically, our model consists of two separate branches that are connected by entropy-based fusion modules depicted on the right. Not only do these modules receive RGB and lidar feature maps as inputs, but also the corresponding entropy maps $\mathcal{E}_{x_i}$ and $\mathcal{E}_{d_i}$ extracted from the RGB image and sparse lidar depth map, respectively. While the motivation for the $max$ operation in the entropy module is explained later in Sec. \ref{Subsec_entropy}, the task of the entropy map after the convolution and the sigmoid activation is to enhance the most informative features while suppressing irrelevant background features \cite{bijelic2019seeing}. We place the entropy modules (green) right after the convolutional layers outside of the bottleneck blocks contained in the feature extractor and the head of YOLO-V3 \cite{Redmon2018YOLOv3AI}. This leads to $6$ fusion modules in the Darknet-53 backbone and another $8$ modules in the corresponding head. Since having a unique branch for each modality increases the number of parameters, we compensate the growth by increasing the number of channels after block $3$ only by $50\%$ resulting in $384$ and $576$ channels, respectively. As will be shown later in the experiment section, we found that this reduced version did not affect the performance for single modality OD. 
\subsection{Leveraging Data Augmentations for Multiple Modalities}
To encourage domain-invariant feature learning and prevent overfitting on the source domain, we leverage data augmentations that mimic the effects of different weather conditions on the sensor. To this end, we notice that each sensor is degraded in a different way. For example, the camera is affected by lighting conditions and fog in a sense that the appearance and color of objects might change, and the contrast of the overall image is reduced. For lidar, the suspended water drops in the air due to fog, rain, or snow result in backscatter, noise, and a lower number of points. To account for the aforementioned effects, we introduce the following augmentations for the lidar depth map:
\begin{itemize}
    \itemsep0em
    \item Points dropout: we choose to drop points from a scene in clear weather conditions with a probability $p_{dropout}$, so the network becomes less sensitive to the sparsity level. In our experiments, $p_{dropout}$ is chosen randomly between $0$ and $0.4$.   
    \item Additive Gaussian noise: to account for uncertainty in the depth measurement, Gaussian noise with a small standard deviation ($1\%$ of the maximum depth value) is added to the depth of each valid point. 
    \item Backscatter points: In snow or rain, some measurements come from the reflection of lidar beams from water droplets. To simulate this effect, we add random points with a probability ($p_{backscatter} = 0.1$), which have a random depth value less than $20\%$ of the maximum depth value. This is based upon the observation that backscatter points are usually closer to the vehicle.  
\end{itemize}

As specific augmentation for RGB images, we rely on color jittering, which applies random changes in hue, saturation and contrast. Further, we use random horizontal flipping, scaling and translation.

\vspace{-0.5em}
\subsection{Entropy-weighted Domain Adversarial Learning} \label{Subsec_entropy}
Domain adversarial training \cite{Ganin2015UnsupervisedDA} seeks to align source and target domain features through a min-max game with a domain classifier. However, enforcing this alignment constraint on all features may result in a suboptimal performance, because some regions in the image and depth map are not transferable between domains. For instance, the background appearance in RGB images varies strongly with different weather and lighting conditions. Similarly, in the lidar depth map, backscattered points should not be aligned. Therefore, considering all features during the alignment may result in a low discriminability of the model on both source and target domains, which is undesirable.

To counteract these negative effects during domain adversarial training, we leverage an existing advantage in our baseline, which is the scaling of features by the sensors' entropy channel. Local measurement entropy assigns a higher value to more uncertain values in the image space, like edges, corners and foreground objects. A multiplication of the deep features with the entropy attenuates the background while giving more weight to foreground features. However, there exists a limitation in the entropy modules of \cite{bijelic2019seeing}: the input feature maps to the detector are taken from the lidar branch, meaning that the lidar's entropy will reweight the fused features which are fed to the detection head. However, the lidar's entropy can deteriorate in adverse weather conditions, when the number of points is significantly reduced, providing less information to the detector about some foreground regions. Moreover, attempting to align the features in the lidar or image branch alone will lead to a bias towards aligning foreground regions determined by the modality's entropy. In contrast, we aim to learn domain-invariant foreground features by exploiting both modality-specific and modality-shared information. To mitigate this negative effect, we modify the entropy fusion scheme of \cite{bijelic2019seeing} and compute the maximum entropy of the lidar and RGB images per pixel $j$: $\mathcal{E}_{max_i}^{j} = max(\mathcal{E}_{x_i}^{j}, \mathcal{E}_{d_i}^{j})$. The resulting entropy map is then used for both streams. Now, $\mathcal{E}_{max_i}$ leverages the most informative regions in the image space that were captured by both sensors. This reduces asymmetric modality noise in different weather conditions.
\begin{figure*}[t!]
    \centering
    \includegraphics[width=1.0\textwidth, height=0.3 \textwidth]{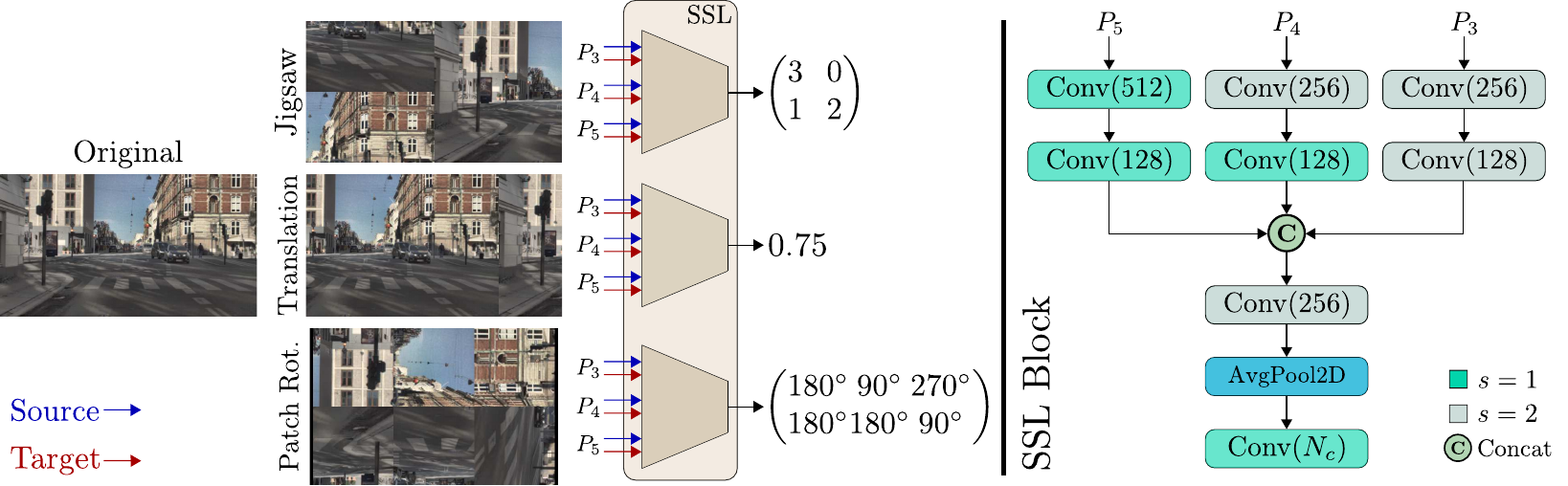}
    \caption{Left: Image examples for each pretext task including the correct label output. Right: One SSL model used to solve one pretext task (s=stride).}
    \label{fig:ssl}
    \vspace{-1em}
\end{figure*}

After enhancing the fusion scheme, we add $3$ domain discriminators $\mathbf{D}_k$ with $k \in \{1,2,3\}$ at $P_3$, $P_4$ and $P_5$ in Fig. \ref{fig:architecture} to align the entropy-weighted instance-level features of different domains at various scales. Each discriminator learns to classify the modality-fused source and target features ($\mathbf{f}_s$ and $\mathbf{f}_t$) into their domain by minimizing the least-squares loss: 
\begin{align}
\label{eq:domain_alignment1}
    \mathcal{L}_{adv_{\mathbf{D}}} &=  \sum_{k=1}^{3} \mathbb{E}\left[(\mathbf{D}_k(\mathbf{f}_t^k))^2\right]
    + \mathbb{E}\left[(1 - \mathbf{D}_k(\mathbf{f}_s^k))^2\right].
\end{align}
To confuse the discriminators, the feature extractor must learn domain-invariant features by minimizing:
 \begin{align}  
    \mathcal{L}_{adv_{\mathbf{F}}} &=  \sum_{k=1}^{3} \mathbb{E}\left[(1 - \mathbf{D}_k(\mathbf{f}_t^k))^2\right].  \label{eq:domain_alignment2}
\end{align}   
Inspired by the GAN framework of \cite{gan}, a feature-matching loss $\mathcal{L}_{fm}$ is added to each discriminator to regularize the training by minimizing the $l_2$-distance between the discriminator source and target features at each layer $l$ in the discriminator. $\mathcal{L}_{fm}$ is defined as follows:
\begin{equation}
\label{eq:feature_matching}
 \begin{aligned}
    \mathcal{L}_{fm} =  \sum_{k=1}^{3}\sum_{l=1}^{L} \left(\mathbf{D}_k^{l}(\mathbf{f}_S^k)- \mathbf{D}_k^{l}(\mathbf{f}_T^k)\right)^2.
\end{aligned}   
\end{equation}

\subsection{Self-supervised learning for DA}
Self-supervised learning is typically used as a pre-training strategy to learn visual representations that generalize well to downstream tasks. In this work, we empirically show that it is not only suitable to align two domains \cite{Sun2019UnsupervisedDA} but can also be very effective in a multi-target unsupervised domain adaptation setting, which is much more important for applications like autonomous driving. 

We begin by redesigning the rotation pretext task \cite{Sun2019UnsupervisedDA}, where a classification head predicts whether an image has been rotated by \{$0^{\circ}$, $90^{\circ}$, $180^{\circ}$, $270^{\circ}$\}. Since we want to promote the learning of instance-level discriminative features, we split an image into $m \times n$ patches, which are then randomly rotated by one of the aforementioned set of angles before being reassembled. Consequently, the output of the pretext model also changes from a single class to a grid of classes, similar to a segmentation output. An example of such a patch rotation including the output is also shown on the left in Fig. \ref{fig:ssl}. We hypothesize that this variant has the advantage, that the classification head depicted on the right may consider instance-level and global features, as inspecting adjacent patches helps to predict the correct class. Note that we apply the same transformation for all modalities and entropy maps, which are not illustrated here. To further encourage the instance-level and global-level feature learning, we use Jigsaw \cite{Noroozi2016UnsupervisedLO}. In this case, the image is reassembled from shuffled patches and the pretext model must predict the correct position of each patch. We slightly modify the output of this task so that it is again a grid. Finally, we also add a translation pretext task, where each modality is rolled in the horizontal direction by a distance $\Delta_x \in \{0, \frac{1}{4}W, \frac{1}{2}W, \frac{3}{4}W\}$, where $W$ is the width in the image space (see Fig. \ref{fig:ssl}). Note that this task is also related to Jigsaw, since a displacement of $\frac{1}{2}W$ can also be realized in Jigsaw for an even number of rows. By combining these tasks, the model can learn the interactions of various regions of the image/ depth map. 

Obviously, the previously mentioned combination of pretext tasks can be easily extended to the multi-target domain adaptation (MTDA) setting, as there is no constraint in SSL to a certain number of domains. However, we observe that SSL under the MTDA is more biased towards domains which contain a larger number of samples. To ensure an equal alignment for all domains, we propose a domain-balanced training for SSL. During each epoch, target domains which have less training examples are oversampled, so that the model observes each domain with the same probability. 

\section{Experiments}
\textbf{Dataset}. DENSE \cite{bijelic2019seeing} is a large-scale autonomous driving dataset captured in middle and northern Europe. It provides data from multiple sensors under adverse weather conditions in real driving scenarios. All frames contain scene tags, indicating the time of day and the weather conditions. 

As we are interested in isolating and studying the effects of different domain discrepancies on the system's performance on a common and general setup involving RGB and lidar, we deviate from \cite{bijelic2019seeing} and use only \textit{Clear Day} as labeled source domain without any night images. For our single-target domain adaptation (STDA) setting, we adapt the model from \textit{Clear Day} to either one of the following splits: \textit{Clear Night}, \textit{Dense Fog Day}, \textit{Light Fog Day}, or \textit{Snow Day}. For the multi-target domain adaptation case, the target domain contains all four aforementioned target splits together. We use the original train, val and test splits ($2183/399/1005$ frames) of \textit{Clear Day}, while splitting the target domain data in the ratio ($75:25$) for train and test, since there are no predefined splits. 
\begin{table*}[t!]
\caption{Single-target domain adaptation results on the test splits of the DENSE dataset.}
\label{table:stda}
\setlength\tabcolsep{5.2pt}
\begin{center}
\scalebox{0.95}{
\begin{tabular}{llccc|ccc|ccc|ccc|ccc}
	\toprule
  & Methods & \multicolumn{3}{c}{Clear Day (*)}  & \multicolumn{3}{c}{* $\rightarrow$ Light Fog Day}  & \multicolumn{3}{c}{* $\rightarrow$ Dense Fog Day}  & \multicolumn{3}{c}{* $\rightarrow$ Snow Day}  & \multicolumn{3}{c}{* $\rightarrow$ Clear Night}  \\
  \midrule
 && Car & Ped. & RV & Car & Ped. & RV & Car & Ped. & RV & Car & Ped. & RV & Car & Ped. & RV\\
  \midrule
\textbf{Unimodal}  & Lidar Only & 57.2 & 45.1 & 34.9 & 51.5 & 14.2 & 0.2 & 27.6 & 20.1 & 8.1 & 51.4 & 43.2 & 9.3 & 61.6 & 50.0 & 20.2  \\
\textbf{without DA} &RGB Only & 79.8 & \textbf{75.5} & 67.7 & 82.2 & 39.2 & 0.2 &  70.9 & 53.3 & \textbf{15.3} & 75.5 & 77.3 & 19.5 & 46.9 & 52.6 & 16.0  \\
 & RGB Only, large & 79.3 & 74.6 & 66.0 & 82.9 & 40.5 & 3.6 & 71.4 & 50.7 & 15.0 & 77.5 & 76.1 & 15.3 & 42.3 & 52.0 & 12.2  \\
\midrule
\textbf{Multimodal} & Entropy Fusion \cite{bijelic2019seeing} & \textbf{81.8} & 74.8 & \textbf{69.8} & 84 & 38.2 & \textbf{5.8} & 71.2 & 50 & 11.3  & 78 & 78.4 & 20.8 &  59.3 & 63.4 & 22.7  \\
\midrule
 & ADDA \cite{ADDA} & 79.8 & 75.5 & 67.7 & 83.5 & 38.6 & 2.5 & 77.3 & 54.5 & 14.8 & 75.1 & 77.2 & 12.7 & 54.3 & 57.8 & 22.5 \\
\textbf{Unimodal DA} & RGB, CycleGAN \cite{Zhu2017UnpairedIT} & 79.8 & 75.5 & 67.7 & 85.5 & 35.2 & 3.7 & 80.2 & 61.3 & 14.2 & 75.6 & 79.1 & 21.5 & 64.0 & 65.6 & 28.3 \\
& CyCADA \cite{CyCADACA} & 79.8 & 75.5 & 67.7 & 82.8 & 33.5 & 1.3 & 78.6 & 55.2 & 15.1 & 74.6 & 77.5 & 19.3 & 63.9 & 66.1 & 29.2 \\
\midrule
\rowcolor[HTML]{EFEFEF}
\textbf{Multimodal DA} &Ours & \textbf{81.8} & 74.8 & \textbf{69.8} & \textbf{85.9} & \textbf{45.0} & 4.6 & \textbf{81.0} & \textbf{63.3} & 14.3 & \textbf{80.8} & \textbf{79.7} & \textbf{26.1} &  \textbf{68.8} & \textbf{67.7} & \textbf{40.5} \\
\bottomrule
\textbf{Oracle} & Entropy Fusion \cite{bijelic2019seeing} & 81.8 & 74.8 & 69.8 & 90.4 & 55.5 & 5 & 90.7 & 77.9 & 15 &  87.2 & 83.3 & 32.1 & 79.0 & 74.1 & 49.7  \\
\bottomrule
\end{tabular}
}
\end{center}
\end{table*}

\textbf{Metrics}.
We evaluate our approach on three classes: Passenger Car (Car), Pedestrian (Ped.), and Ridable Vehicle (RV), using the KITTI evaluation framework \cite{kitti}. For Car, we report the Average Precision with an overlap of $70\%$, AP70, while for Pedestrian and RV we report AP50. Each class is evaluated on three difficulty levels (easy, moderate and hard) and the mean per class is reported in the tables. 

\textbf{Implementation Details}.
We implement our method on top of the YOLO-V3 repository \footnote{\url{https://github.com/ultralytics/yolov3}}. In all experiments, we use a batch size of 10, an SGD optimizer with a learning rate of 0.013 for the detection model and an ADAM optimizer with a learning rate of 0.0001 for the discriminators $\mathbf{D}_k$ and the SSL models. We train our network in two steps. First, we perform an adversarial warm-up in which we minimize the detection loss as well as the adversarial loss. Then, we stop the adversarial learning and replace it with self-supervised learning for both source and target data. 

\textbf{Baselines}. For the STDA setting, we evaluate our approach against several baseline methods. These include the two unimodal non domain adaptation settings, where we train the model only on \textit{Clear Day}, using either the RGB branch (RGB Only) or the lidar branch (Lidar Only) of our network. As we have reduced the number of parameters to compensate for the multiple branches, we further consider using the same number of channels as in the original YOLO-V3 network, which we denote as (RGB Only, large). In this setting, we also evaluate the multimodal entropy fusion approach of \cite{bijelic2019seeing}, which does not use the maximum entropy during fusion. For the experiments involving domain adaptation, we consider the unimodal approaches ADDA \cite{ADDA}, CyCADA \cite{CyCADACA}, and training with additional, style transferred RGB images (RGB, CycleGAN). Finally, we also include the upper bound for our multimodal baseline which uses target labels during training (Oracle).

\begin{table*}[h]
\label{table:mtda}
\caption{Multi-target domain adaptation results on the test splits of the DENSE dataset. The mean in the last column is calculated on the $4$ target domains on all classes.}
\setlength\tabcolsep{5.9pt}
\begin{center}
\scalebox{1.0}{
\begin{tabular}{lccc|ccc|ccc|ccc|ccc|c}
	\toprule
  Configuration & \multicolumn{3}{c|}{Clear Day}  & \multicolumn{3}{c}{Light Fog Day}  & \multicolumn{3}{c}{Dense Fog Day}  & \multicolumn{3}{c}{Snow Day}  & \multicolumn{3}{c|}{Clear Night} & \multirow{2}{*}{Mean}\\
  \cmidrule{1-16}
  & Car & Ped. & RV & Car & Ped. & RV & Car & Ped. & RV& Car & Ped. & RV& Car & Ped. & RV &   \\
  \midrule
  
\textbf{A} Entropy Fusion \cite{bijelic2019seeing} & 81.8 & 74.8 & 69.8 & 84.0 & 38.2 & 5.8 & 71.2 & 50.0 & 11.3  & 78.0 & 78.4 & 20.8 &  59.3 & 63.4 & 22.7 & 48.0  \\  
\textbf{B} + Max Entropy & 81.6 & 75.3 & 69.3 & 85.1 & 40.6 & 5.0 & 72.2 & 55.0 & \textbf{14.8} & 78.4 & 76.9 & 13.1 & 59.3 & 62.8 & 27.0 & 48.4  \\
\textbf{C} + Augmentations & 82.5 & 75.9 & 69.6 & 82.7 & 45.2 &\textbf{5.8} & 72.7 & 58.6 & 13.7 & 78.6 & \textbf{78.7} & 13.2 & 57.6 & 62.4 & 25.3 & 48.5  \\
\textbf{D} + Discriminator & 81.0 & 74.0 & 69.4 & 85.4 & 42.6 & 1.6 & \textbf{81.7} & 60.1 & 14.4 & 79.1 & 77.5 & 21.1 & 65.5 & 65.9 & 30.5 & 51.8 \\

\textbf{E} + SSL & \textbf{82.7} & 76.2 & \textbf{74.4} & 86.1 & 44.4 & 3.3 & 76.6 & 59.3 & 13.7 & \textbf{81.0} & \textbf{78.7} & \textbf{21.5} & \textbf{68.8} & \textbf{67.6} & 34.8 & 53.2
 \\
 \rowcolor[HTML]{EFEFEF}
 \textbf{F} + Domain Balanced  & \textbf{82.7} & \textbf{76.3} & 72.3 & \textbf{86.8} & \textbf{51.0} & 2.7 & 77.4 & \textbf{61.8} & 14.3 & 80.8 & 78.5 & 17.6 & 66.6 & 66.4 & \textbf{37.5} & \textbf{53.5}
 \\
\bottomrule
Oracle & 84.6 & 76.8 & 74.3 & 93.0 & 52.5 & 4.0 & 92.8 & 74.6 & 15.0 &  87.0 & 82.0 & 29.6 & 81.2 & 74.7 & 49.3 & 61.7 \\
\bottomrule
\end{tabular}
}

\end{center}
\end{table*}
\section{Results}
\textbf{STDA Results.} The results of our STDA setting are shown in Table~\ref{table:stda}. First, it can be seen that a smaller unimodal network (RGB Only) does not degrade performance compared to a twice as large version (RGB Only, large) when trained on source data only. Please note that the results for different target domains can increase relative to the source domain, due to varying scene complexities \cite{bijelic2019seeing} and a smaller number of instances per image (for example there are less pedestrians and ridable vehicles in snow/rain). Next, we validate the effectiveness of the deep-entropy fusion module from \cite{bijelic2019seeing} which outperforms both RGB Only and Lidar Only on most splits. An exception for this is the dense fog domain, where RGB Only clearly surpasses the fusion model on two out of three classes. We attribute this to the fact that lidar exhibits a stronger degradation in dense fog and that the non-maximum entropy fusion mechanism may be suboptimal here. On the other hand, we can see that using lidar-only is better than RGB-only on the \textit{Clear Day} $\rightarrow$ \textit{Clear Night} domain gap, which is in line with previous works \cite{bijelic2019seeing, 8885465}, and has its reason in the simple fact that lidar is not degraded at all during \textit{Clear Night} in contrast to RGB camera. 

To study the impact of each domain change, we further show in Table \ref{table:stda} the upper bound (Oracle) where an entropy fusion model was trained in a supervised manner on both source and target data. Clearly, the domain gap is especially large in the domains \textit{Dense Fog Day} and \textit{Clear Night} and also for smaller classes, like Pedestrian or Ridable Vehicle, which are vulnerable road users and should therefore be detected. We also observe that some unimodal DA methods are able to outperform the entropy fusion mechanism, which further motivates the need for multimodal domain adaptation. 

If we now apply our proposed multimodal domain adaptation method, we can further reduce the model's performance gap compared to the upper bound. For instance, $50\%$ of the domain gap is bridged for the class Car in \textit{Dense Fog Day} ($9.8$ AP points recovered from a domain gap of $19.5$ points), $48\%$ for Pedestrian in \textit{Dense Fog Day} and up to $66\%$ for the RV class in \textit{Clear Night}. The relative gains are also visible in domains like \textit{Light Fog Day} and \textit{Snow Day}, even when the domain gap is smaller.
\begin{table}[t!]
\caption{Ablation study on the discriminator and its input for Clear Day $\rightarrow$ Dense Fog Day. Conducted before SSL training.}
\label{table:ablation_disc}
\setlength\tabcolsep{5.0pt}
\begin{center}
\scalebox{0.84}{
\begin{tabular}{lccc>{\columncolor[gray]{0.95}}c|ccc>{\columncolor[gray]{0.95}}c}
	\toprule
  & \multicolumn{4}{c}{Clear Day}  & \multicolumn{4}{c}{Dense Fog Day}  \\
  \midrule
  Method & Car & Ped. & RV & Mean & Car & Ped. & RV & Mean \\
  \midrule
  Entropy Fusion \cite{bijelic2019seeing} & \textbf{81.8} & \textbf{74.8} & \textbf{69.8} & \textbf{75.5} & 71.2 & 50 & 11.3  & 44.2 \\
from RGB branch  & 78.9 & 72.7 & 64.8 & 72.1 & 76.2 & \textbf{57.6} & 14.4 & 49.4 \\  
from Lidar branch & 78.5 & 71.7 & 63.7 & 71.3 & 79.5 & 55.9 & 13.7 & 49.7\\  
\rowcolor[HTML]{EFEFEF}
MaxEntropy & 80.1 & 73.5 & 66.5 & 73.4 & \textbf{82.9} & 54.3 & \textbf{15} & \textbf{50.7} \\  
\midrule
MaxEntropy - no $\mathcal{L}_{fm}$ & 79.7 & 71.9. & 68.4 & 73.3 & 80 & 53.7 & 13.5 & 49.1 \\  
\bottomrule
\end{tabular}
}

\end{center}
\end{table}
\begin{table}[t]
\caption{Ablation study on various pre-text tasks for Clear Day $\rightarrow$ Dense Fog Day.}
\label{table:ablation_ssl}
\setlength\tabcolsep{4.5pt}
\begin{center}
\scalebox{0.85}{
\begin{tabular}{lccc>{\columncolor[gray]{0.95}}c|ccc>{\columncolor[gray]{0.95}}c}
	\toprule
   & \multicolumn{4}{c}{Clear Day}  & \multicolumn{4}{c}{Dense Fog Day}  \\
  \midrule
  Method & Car & Ped. & RV & Mean & Car & Ped. & RV & Mean \\
  \midrule
Entropy Fusion \cite{bijelic2019seeing} & \textbf{81.8} & \textbf{74.8} & \textbf{69.8} & \textbf{75.5} & 71.2 & 50 & 11.3  & 44.2 \\
Rotation \cite{Misra2020SelfSupervisedLO}  & 78.8 & 73.9 & 67.3 & 73.3 & 75.3 & 54.2 & 13.5 & 47.7 \\  
Patch Rotation & 80.4 & 74.4 & 65.6 & 73.5 & 74.8 & 59.7 & 14.3 & 49.6 \\
Jigsaw \cite{Noroozi2016UnsupervisedLO} & 78.8 & 73.2 & 64.8 & 72.3  & 74.7 & 59.9 & 13.5 & 49.4 \\
Translation \cite{Misra2020SelfSupervisedLO} & 79.8 & 73.0 & 65.4 & 72.7 & 79.9 & 55 & \textbf{15} & 50.0 \\  
\rowcolor[HTML]{EFEFEF}
Patch Rot. + Jig. + Trans. & 80.1 & 73.7 & 65.2 & 73.0 & \textbf{80.0} & \textbf{61.7} & 13.6 & \textbf{51.8} \\
\bottomrule
\end{tabular}
}
\vspace{-1em}
\end{center}
\end{table}

\textbf{MTDA Results.} In Table~\ref{table:mtda}, we evaluate our methodology under the MTDA setting, which is arguably more relevant for real-world deployment in autonomous vehicles, since unlabeled data in multiple domains is often easy to collect and a good performance on all these domains at once is required. Moreover, performing multiple pairwise domain adaptations sequentially might lead to catastrophic forgetting \cite{mccloskey1989catastrophic} on previous domains \cite{bobu2018adapting}, which is undesirable. In this setting, we start by showing the results of the baseline in configuration A. Then, we demonstrate how cumulative effects increase the performance. In configuration B, our proposed maximum entropy fusion mechanism is employed and is shown to further improve the fusion, surpassing the results of RGB-only in Table~\ref{table:stda}. However, there is a drop in performance on the Ridable Vehicle class in the domain \textit{Snow Day}. The modality-specific augmentations (configuration C) again improve the results, especially in the light and dense fog domains, since the lidar augmentations can simulate to an extent the adversary effects of these two domains. Next, we add the multi-scale discriminators in configuration D and notice a substantial overall increase for most of the target domains, albeit with a slight drop in the source domain. Note that we do not extend the domain discriminators to classify multiple domains. Instead, they only predict either a source or target class. We hypothesize that predicting multiple domains is not necessary for the discriminator since the maximum entropy is able to re-weight the transferable and non-transferable parts of the image space for the discriminator. This result is not conclusive and more investigations into the discriminator classification loss are needed. In configuration E, we train the network in a second stage on the pretext tasks and remark an overall increase in the AP, with one exception in the \textit{Dense Fog Day} split. We attribute this discrepancy to the lower number of samples contained the \textit{Dense Fog Day} split compared to others. The domain-balanced training scheme (configuration $F$) enhances the performance on both the \textit{Dense Fog Day} and \textit{Light Fog Day} splits, with an overall increase in the mean performance. Note that in this set of experiments, the oracle is calculated by joint supervised training on all $4$ domains, whereas the oracle in the STDA experiments was calculated by joint training on the source domain and one target domain at a time. Taking a look at the last column in Table~\ref{table:mtda}, the mean performance of the proposed multimodal, multi-target DA outperforms the baseline (configuration A) by a significant margin ($40\%$ of the domain gap). Besides, the proposed approach has the effect of improving the baseline performance on the source domain as well. 


\textbf{Ablations study on adversarial learning.} We investigate some design choices for the multi-scale discriminators on the \textit{Clear Day} $\rightarrow$ \textit{Dense Fog Day} task and report the results in Table~\ref{table:ablation_disc}. Note that these results were obtained after finishing the adversarial warm-up phase, which precedes the SSL training. First, we abandon our maximum entropy strategy and examine the effect, when the inputs for the discriminators come from either the lidar or RGB branch. As can be seen, without our maximum entropy strategy, both setups lead to suboptimal results on the source domain. However, there is still a significant performance increase on the the target domain, especially when the lidar features are aligned, because in dense fog, the lidar exhibits a stronger degradation than the image. Now, if we use the features of the lidar branch in combination with the maximum entropy fusion, one can see an increased transferability and discriminability. Last, we investigate our setup without using the feature matching loss, and we observe a large performance drop on the target domain, confirming that it stabilizes the discriminator training. 

\textbf{Ablation study on the pretext tasks.} In Table~\ref{table:ablation_ssl}, we study the effect of various pretext tasks for \textit{Clear Day} $\rightarrow$ \textit{Dense Fog Day}. This time, the experiments are conducted without the adversarial warm-up or the maximum entropy fusion. We compare our proposed patch rotation task with the standard rotation task in \cite{Misra2020SelfSupervisedLO}, where a square patch is randomly cropped from the whole image and randomly rotated. Compared to the standard rotation, the patch rotation has the advantage of working at a finer level because the network has to consider the interactions between patches to predict the rotation of each patch, resulting in a higher AP on the \textit{Dense Fog Day} split. Moreover, the patch rotation task allows the network to observe more data compared to the standard rotation task, where the performance depends on which crops are seen during each epoch. In our experiments, we divide the image into $2 \times 3$ patches, while in the jigsaw task, we divide it into $2 \times 2$. While the jigsaw and the translation task perform comparable and slightly better on the target domain, they lag on the source domain. By stacking multiple pretext tasks, the network learns more useful visual cues that refine the detection on the target domain.

\section{Conclusion}
In this work, we proposed an unsupervised domain adaptation approach to improve the performance of a multimodal object detector on a single and multiple target domains. Our approach combines a data augmentation scheme that simulates the effects of different weather conditions, an enhanced entropy fusion strategy that extracts complementary information, and a domain alignment scheme through adversarial and self-supervised learning. A multi-scale discriminator that aligns the instance-level features of a YOLO-V3 based network was introduced along with a feature matching loss. Moreover, we choose and redesign pretext tasks that provide a supervisory signal to the feature extractor, which in turn aligns the domains without majorly sacrificing discriminability on the source domain. We have also shown that the proposed framework can work under a multi-target domain adaptation settings, where we have reframed self-supervised learning to be a powerful tool to achieve this goal. We believe the proposed approach can have an impact on real-world deployment of autonomous vehicles and robotics, and will encourage further investigations in the role of SSL on multi-target multimodal UDA.  


\bibliographystyle{IEEEtran}
\bibliography{IEEEabrv,refs}

\end{document}